 \documentclass[pmlr,twocolumn]{jmlr} 

\usepackage{booktabs}
\usepackage[load-configurations=version-1]{siunitx} 
 
\usepackage{amsmath}
\usepackage{adjustbox}
\usepackage{array}
\usepackage{siunitx} 
\usepackage{xcolor}
\usepackage[normalem]{ulem}
\usepackage{titlesec}
\usepackage{multirow}
\usepackage{arydshln}
\usepackage{todo}

\theorembodyfont{\upshape}
\theoremheaderfont{\scshape}
\theorempostheader{:}
\theoremsep{\newline}

\jmlrvolume{ML4H Extended Abstract Arxiv Index}
\jmlryear{2020}
\jmlrsubmitted{2020}
\jmlrpublished{}
\jmlrworkshop{Machine Learning for Health (ML4H) 2020}

\title[Augmenting BERT Carefully with Underrepresented Linguistic Features]{Augmenting BERT Carefully \linebreak with Underrepresented Linguistic Features}

  \author{}

\author{%
\Name{Aparna Balagopalan} \Email{aparna@winterlightlabs.com}\\
\addr Winterlight Labs, University of Toronto, Vector Institute \\
\Name{Jekaterina Novikova} \Email{jekaterina@winterlightlabs.com}\\
\addr Winterlight Labs
}

\begin{document}

\maketitle

\begin{abstract}
Fine-tuned Bidirectional Encoder Representations from Transformers (BERT)-based sequence classification models have proven to be effective for detecting Alzheimer's Disease (AD) from transcripts of human speech. However, previous research shows it is possible to improve BERT's performance on various tasks by augmenting the model with additional information. In this work, we use probing tasks as introspection techniques to identify linguistic information not well-represented in various layers of BERT, but important for the AD detection task. We supplement these linguistic features in which representations from BERT are found to be insufficient with hand-crafted features externally, and show that jointly fine-tuning BERT in combination with these features improves the performance of AD classification by upto 5\% over fine-tuned BERT alone.\end{abstract}
\begin{keywords}
Alzheimers disease, dementia detection, BERT,  feature  engineering,  transfer learning
\end{keywords}

\section{Introduction}
BERT (Bidirectional   Encoder   Representations from Transformers)~\citep{devlin2019bert} builds on Transformer networks to pre-train bidirectional representations of text by conditioning on both left and right contexts jointly in all layers. These models have achieved state-of-the-art on a variety of tasks in NLP, when fine-tuned. Previous research~\citep{jawahar2019does} has hence suggested that it encodes language information (lexics, syntax etc.) that is known to be important for performing complex natural language tasks.

The task we focus on in this work is Alzheimer’s  disease  (AD) detection from transcripts of speech. AD is a  neurodegenerative disease affecting  over  40  million  people  worldwide~\citep{prince2016world}. Previous research has indicated that spontaneous speech elicited using pictures can be used to detect AD and serve as a quick, objective AD assessment~\citep{fraser2016linguistic}.   
Several  studies  have  used  ML-based speech analysis  to  distinguish  between  healthy  and  cognitively  impaired  speech  of participants in picture description datasets~\citep{balagopalan2018effect,zhu2019detecting}. Clinical literature in the space of AD has identified specific linguistic features, such as part-of-speech (POS) tag frequencies and measures of lexical diversity~\citep{bucks2000analysis} particularly indicative of AD.  Past research has also shown that fine-tuned BERT can be used for AD detection~\citep{balagopalan2020bert}. 

Diagnostic probing tasks~\citep{adi2016fine,conneau2018you} are useful in trying to understand language phenomena encoded by high dimensional embeddings, and have become standard in studying properties of high-level representations. The probing setup consists of using representations to predict a linguistic property of interest. As such, probing tasks are a useful tool to understand what linguistic features are under-represented in BERT. 

~\cite{jawahar2019does} used probing tasks to show that the  intermediate layers in BERT encode a rich hierarchy of linguistic information, and showed that pre-training BERT on a variety of auxilary tasks, as proposed in ~\cite{devlin2019bert}, improves the predictive performance and hence the language information encoded. We attempt to extend this work by using diagnostic tasks proposed by ~\cite{conneau2018you} to probe representations from a BERT model fine-tuned for AD detection.

Prior research has also shown that neural language representation models can be supplemented with domain-specific knowledge either while training~\citep{cai2019multi}, or fine-tuning~\citep{wang2020combining}. 
Motivated by this, we study if BERT can be augmented with additional linguistic features relevant to the AD detection task. Probing tasks are used as introspection techniques to study the linguistic information encoded in BERT, and identify the list of linguistic features to supplement BERT externally with.

The main contributions of our paper are:
(1) We benchmark fine-tuned BERT models at utterance-level on a standard dataset for AD detection from picture descriptions, (2) We probe linguistic information encoded in BERT models fine-tuned for AD detection using diagnostic tasks proposed by ~\cite{conneau2018you}, (3) We augment BERT by combining representations from pre-trained BERT and manually extracted linguistic features, where the additional features represent the linguistic information only present to a small degree in BERT representations. We show that performance of the augmented model increases by 5\% comparing to a non-augmented fine-tuned BERT model.

\section{Related Work}
\paragraph{BERT for AD Detection:}
Bidirectional  Encoder  Representations from  Transformers (BERT)~\citep{devlin2019bert} is a deep neural language representation model. Past research has shown that BERT can be used for AD detection~\citep{balagopalan2020bert}.
Prior work has shown that pre-trained BERT models do not encode several phenonmena related to lexics, syntax, and semantics~\citep{jawahar2019does}. Since clinical literature shows that AD detection from transcribed speech highly depends on a specific set of linguistic features, it is important to study, and potentially enhance the linguistic properties that a BERT-model fine-tuned for an AD detection task encompasses. 

\paragraph{Augmenting Neural Language Representation Models:}
Incorporating domain-specific external knowledge in neural language representations is a field of research that has been actively explored~\citep{cai2019multi}. ~\cite{wang2020combining} combined fine-tuning with a feature-based approach for aspect extraction. Similarly, ~\cite{cai2019multi} combined domain specific word-embeddings with general word embeddings for performance boosts in a sentiment classification task.  ~\cite{kiela2018dynamic} developed methods to dynamically combine neural embeddings from different sources and obtain robust sentence embeddings. 

\paragraph{Probing tasks:}
\label{sec:probing}
Diagnostic tasks~\citep{hupkes2018visualisation, conneau2018you}  reate auxiliary classification tasks for representations, by using the embedded representations to predict a specific linguistic feature, such as predicting the length of a sentence from its sentence-level representation. The intuition is that if the classifier trained using these representations are capable of predicting the linguistic feature well, then it likely encodes information about the linguistic phenomenon. In this work, we use probing tasks following ~\citep{jawahar2019does} and ~\citep{conneau2018you} to assess linguistic features captured in various layers of a BERT model fine-tuned for an AD detection downstream task. We use 5 probing tasks (see Table~\ref{tab:probing}), focusing on surface and syntactic phenomenon.

\vspace{-1em}
\section{Dataset}
For all our experiments, we use DementiaBank~\citep{becker1994natural}, which is a longitudinal dataset of speech for assessing cognitive impairment. It contains 473 narrative picture descriptions~\citep{becker1994natural} where each participant describes a picture shown to them.
We divided each transcript into individual utterances. We treat each utterance as a sample similar to methodology followed by ~\cite{karlekar2018detecting}. In contrast to ~\cite{karlekar2018detecting} we only utilize data from these picture description task. This is done because other speech tasks are exclusively performed by participants with AD. Hence, the aim of our classification setup is of AD detection at utterance level from transcripts of speech elicited via pictures. 
We benchmark all classification models at utterance level, i.e, each sentence spoken by the participant is associated with the diagnosis label of the participant. On performing this, our sample size increases to 5103 utterances, out of which we use about 82\%/9\%/9\%  for train/development/testing respectively, similar to ~\cite{karlekar2018detecting}. See Table~\ref{tab:db_split} in Appendix for the exact split details.

\section{Probing Intermediate Representations of Fine-tuned BERT}
\label{sec:probing}
We fine-tune BERT on DementiaBank for the AD detection task, and probe representations of the first classification token (`[CLS]') from each layer~\citep{jawahar2019does}. We train Multi-layer Perceptrons (MLPs) using embedded representations from various layers of BERT as input to predict the following 5 properties:

\textbf{WordContent}: Given a (word, sentence) pair, predict if the word is present in the sentence or not.

\textbf{SentenceLength}: Given a sentence, predict its length in number of word tokens.

\textbf{TopConstituents}: Given a sentence, predict the sequence  of  top-level constituents in its syntax tree.

\textbf{TreeDepth}: Given a sentence, predict the length of its syntactic parse-tree.

\textbf{BiGramShift}: Given a sentence, predict whether adjust words are inverted or word order is preserved (For example, inversion is seen ``This an is example sentence.").
Grid-search hyperparameter optimization is performed to arrive at the optimal parameter setting using the validation set.
We observe that performance on the syntactic task of tree-depth prediction is low, as is the performance on the word content prediction task (see Table~\ref{tab:probing}). 

Prior work has shown that features relying on both of the above underrepresented properties are important for AD detection from picture descriptions~\citep{fraser2016linguistic, yancheva2015using}. Particularly, variations in proportions of various production rules from the constituency parse representation, depth of the constituency parse tree, etc., which are features of syntactic type, were mentioned to be an important characteristic of impaired speech in~\cite{yancheva2015using}. Presence of informative content words such as ``cookie" or ``boy" while describing the picture, was mentioned as an important characteristic in \cite{fraser2016linguistic} and it is associated with the features emphasized by the word content prediction probing task. Note that these important content words for the picture stimulus have been identified by clinicians, and not arbitrarily defined (see Appendix~\ref{app:info_units}).

\begin{table}[t]
\begin{center}
\begin{adjustbox}{max width=\linewidth}
\begin{tabular}{ |c|c|c|c|  } 
    \hline 
     \textbf{Linguistic Feature} & \textbf{Highest Accuracy} & \textbf{Layer} & \textbf{Feature Type} \\
     \hline
     \textbf{WordContent} & $\textbf{22.47}$ & $\textbf{4}$ & \textbf{Surface} \\
     SentenceLength & $92.81$ & $3$ & Surface \\
     TopConstituents & $80.86$ & $7$ & Syntactic \\
     \textbf{TreeDepth} & $\textbf{36.14}$ & $\textbf{6}$ & \textbf{Syntactic} \\
     BiGramShift & $85.42$ & $12$ & Syntactic \\
     \hline
\end{tabular}
\end{adjustbox}
\caption{\footnotesize{Probing results on BERT fine-tuned on AD classification task. Bold indicates the \textbf{worst} performance in each feature type.}}
\label{tab:probing}
\end{center}
\end{table}

\section{Classification Results}
Based on our observations in Section~\ref{sec:probing}, we identify that externally hand-engineered features capturing: (1) presence of informative content words in utterances, and (2) syntactic tree depths, might help in improving the AD detection performance when combined with BERT. We extract 119 -- 117 syntactic and 2 word-content based -- features at utterance-level. The two word-content features we extract are: (1) a boolean indicating the presence of informative content units (see Appendix~\ref{app:features} for list), (2) total number of information content units in each utterance. The 117 syntactic features include depth-related features of the constituency parse representations, the height of the constituency parse-tree, proportion of verb-phrases, proportion of production rules of type ``adjective phrase followed by adjective", etc.

We compare several input settings to see the effect of these features:

\textbf{NN with FS1:}  A Neural Network (NN) classifier using the set of all 119 features 
    
\textbf{Fine-tuned BERT:} Fine-tuning a BERT sequence classification model, where a linear layer maps the concatenation of the final hidden layer representation from BERT to binary class labels~\citep{wolf2019huggingface}.
    
\textbf{ 
    BERT+FS1}: Fine-tuning a BERT sequence classification model, where a linear layer maps the concatenation of the final hidden layer representation from BERT and the feature vector to  binary class labels. 

All classification models are optimized to the best possible parameter setting with 5-fold grid-search cross-validation (see Appendix~\ref{app:hyperparam}).

\begin{table}[t]
\begin{center}
\begin{adjustbox}{max width=\linewidth}
\begin{tabular}{ |c|c|c|c|  } 
    \hline 
     \textbf{Model} & \textbf{Accuracy} & \textbf{Sensitivity} & \textbf{Specificity} \\
     \hline
     NN + FS1& $0.63$ &$\textbf{0.64}$ &  $0.62$\\
     Fine-tuned BERT & $0.71$& $0.62$& $0.79$\\
     BERT + FS1 & $\textbf{0.76}$& $0.63$ & $\textbf{0.86}$\\
     \hline
    \end{tabular}
\end{adjustbox}
\label{tab:2}
\end{center}
\caption{\footnotesize{Results on the AD detection task with (1) NN using FS1, (2) Fine-tuned BERT using text, (3) Augmented BERT using text and FS1. FS1 is the feature set identified via linguistic probing in Table~\ref{tab:probing}\label{tab:clf}. Bold indicates \textbf{highest} performance.
}}
\end{table}

We find that fine-tuned BERT models are able to perform well above chance for the AD detection task, similarly to previous findings by ~\cite{pompili2020inesc,balagopalan2020bert}.
We observe that the third setting (BERT+FS1) attains the highest accuracy, which is about 13\% accuracy points higher than using classification models using features alone, and about 5\% higher than fine-tuning BERT alone (see Table~\ref{tab:clf}). 

\section{Discussion}
In this work, we probe intermediate representations of fine-tuned BERT models for an Alzheimer's Disease task to understand its deficiencies. We identify that there is scope for better representation of two categories of features important for the downstream task --- (1) constituency tree-based features, (2) presence of specific informative words in the text. On supplementing these features by engineering them separately, and jointly fine-tuning them with BERT, we achieve a 5\% increase in predictive performance. This is promising for reliably detecting AD early.
Though we use probing tasks~\citep{pimentel-etal-2020-information, conneau2018you} exclusively to identify scope of improvement in BERT models, this can be replaced with other methods, such as generating explanations~\citep{mothilal2020explaining}. Ongoing work is on performing similar analyses with other neural language representation models.


\bibliography{jmlr-sample}

\clearpage

\appendix
\section{DementiaBank}
The table below shows the train/val/test splits of DementiaBank we benchmark all models settings ons.
\begin{table}[ht!]
    \centering
    \begin{tabular}{ |c|c|  } 
    \hline 
     \textbf{Data subset} & \textbf{\# utterances} \\
     \hline
     Train & 4269\\
     Validation & 429\\
     Test & 409 \\
     \hline
    \end{tabular}
    \caption{\footnotesize{DementiaBank statistics\label{tab:db_split}}}

\end{table}

\section{Feature list}
\label{app:features}
Table~\ref{tab:lex_features} contains a summary of all the engineered linguistic features we extract.
\begin{table*}[t!]
\centering
\caption{Summary of all underrepresented features extracted. The number of features in each subtype is shown in the second column (titled "\#features").\label{tab:lex_features}
}
{
\begin{adjustbox}{max width=\linewidth}

\begin{tabular}{lcc}
 \textbf{Feature type} & \textbf{\#Features} & \textbf{Brief Description} \\
 \toprule
Production Rules and parse tree depth &104 & Number of times a production type occurs divided by total number of productions \\
\midrule
Phrasal type ratios  & 13 & Proportion, average length and rate of phrase types\\
\midrule
Word Content & 2 & Presence of information content units in utterance \\
\bottomrule
\end{tabular}
\end{adjustbox}
}
\end{table*}

\section{Hyperparameter Tuning}
\label{app:hyperparam}
We search for the optimal set of parameters by grid-search based on the validation performance. 

\subsection{Finetuning BERT}
The search space for finetuning BERT was:
\begin{itemize}
    \item number of epochs $\in$ $\{2,3,4,5,6\}$, and
    \item Adam initial learning rate $\in$ \{\num{2e-5}, \num{2e-4}\}.
\end{itemize}

\subsection{NN + FS1}
The search space for the parameters of the NN model was:
\begin{itemize}
    \item layers $\in$ $\{1,2,3\}$
    \item number of units per layer $\in$ $\{10,100\}$.
\end{itemize}

\section{Information Content Units}
\label{app:info_units}
The list is: ``boy", ``son", ``brother", ``girl", ``daughter", ``sister", ``female", ``woman", ``adult", ``grownup", ``mother", ``lady",
                         ``cookie", ``biscuit", ``treat", ``cupboard", ``closet", ``shelf", ``curtain", ``drape", ``drapery", ``dish", ``cup", ``counter", ``apron", ``dishcloth", ``dishrag", ``towel", ``rag", ``cloth",
                         ``jar", ``container", ``plate", ``sink", ``basin", ``washbasin", ``washbowl", ``washstand", ``tap", ``faucet", ``stool", ``seat", ``chair",
                                             ``water", ``dishwater", ``liquid", ``window", ``frame", ``glass", ``floor", ``outside", ``yard", ``outdoors", ``backyard", ``garden",
                                             ``driveway", ``path", ``tree", ``bush", ``exterior", ``kitchen", ``room", ``take", ``steal", ``fall", ``ignore", ``notice", ``daydream", ``pay",
                                             ``overflow", ``spill", ``wash", ``dry", ``sit", ``stand"
\end{document}